\documentclass{article}

\usepackage[preprint]{neurips_2025}

\usepackage[utf8]{inputenc} 
\usepackage[T1]{fontenc}    
\usepackage{hyperref}       
\usepackage{url}            
\usepackage{booktabs}       
\usepackage{amsfonts}       
\usepackage{nicefrac}       
\usepackage{microtype}      
\usepackage{xcolor}         
\usepackage{graphicx}
\usepackage{amsmath}

\title{KHRONOS: a Kernel-Based Neural Architecture for Rapid, Resource-Efficient Scientific Computation}

\author{
  Reza T. Batley\\
  Kevin T. Crofton Department of Aerospace and Ocean Engineering\\
  Virginia Polytechnic Institue and State University\\
  Blacksburg, VA 24060 \\
  \texttt{rezabatley@vt.edu} \\
  \And
  Sourav Saha \\
  Kevin T. Crofton Department of Aerospace and Ocean Engineering\\
  Virginia Polytechnic Institue and State University\\
  Blacksburg, VA 24060 \\
  \texttt{souravsaha@vt.edu} \\
}

\begin{document}

\maketitle

\begin{abstract}
Contemporary models of high-dimensional physical systems are constrained by the curse of dimensionality and a reliance on dense data. We introduce KHRONOS (Kernel-Expansion Hierarchy for Reduced-Order, Neural-Optimized Surrogates), an AI framework for model-based, model-free and model-inversion tasks. KHRONOS constructs continuously differentiable target fields with a hierarchical composition of per-dimension kernel expansions, which are tensorized into modes and then superposed. We evaluate KHRONOS on a canonical 2D, Poisson equation benchmark: across 16-512 degrees of freedom (DoFs), it obtained $L_2^2$ errors of $5\times 10^{-4}$ down to $6\times 10^{-11}$. This represents a $> 100\times$ gain over Kolmogorov-Arnold Networks (which itself reports a $\sim 100\times$ improvement on MLPs/PINNs with $100\times$ fewer parameters) when controlling for the number of parameters. This also represents a $\sim 10^{6}\times$ improvement in $L_2^2$ error compared to standard linear FEM at comparable DoFs. Inference complexity is dominated by inner products, yielding sub-millisecond full-field predictions that scale to an arbitrary resolution. For inverse problems, KHRONOS facilitates rapid, iterative level set recovery in only a few forward evaluations, with sub-microsecond per-sample latency. KHRONOS’s scalability, expressivity, and interpretability open new avenues in constrained edge computing, online control, computer vision, and beyond.
\end{abstract}

\section{Introduction}
Since Rosenblatt's \textit{perceptron} \cite{rosenblatt1958perceptron}, \textit{multilayer perceptrons} (MLPs) or \textit{artificial neural networks} have come a long way in both data-driven and scientific modeling \cite{jordan2015machine,cuomo2022scientific,brunton2024promising}. Many variations of neural networks have been proposed to achieve specific goals \cite{lecun1989backpropagation,rumelhart1985learning,vaswani2017attention,raissi2019physics}. However, at their core, most network architectures have remained the same; passing data through a set of activation functions, multiplying outputs by weights and biases to construct a non-linear mapping from input to output. Despite tremendous success, traditional neural architectures suffer from the curse of dimensionality: an exponential growth of trainable parameters for very high-dimensional and complex problems. This has helped lead to a six-order-of-magnitude increase in the cost of training from 2012 to 2018 \cite{schwartz2020green}. In addition, interpretability and transferability remain a significant challenge for traditional neural networks. 

Consequently, alternative network structures and activation functions have been explored, including kernel function-based non-parametric activations \cite{scardapane2019kafnets}. Some works have discussed, at length, the mathematics of kernel-based activation functions \cite{huang2014kernel,ghorbani2020neural,seleznova2022neural}. However, these works do not discuss how to reduce the size of the network while maintaining accuracy. Recently, the proposition of Kolmogorov-Arnold Neural Networks (KAN) provided a fresh perspective on neural architecture \cite{liu2024kan}. KANs embed basis functions, including kernel functions, into the data space instead of so-called neurons. KANs have shown impressive performance in many applications, demonstrating the promise of such alternative architectures \cite{ranasinghe2024ginn,liu2024kan,thakolkaran2025can}. Despite that, KANs still follow a collocation-based sampling method that leaves room for further reduction in structure and improvement in performance.         

In this work, we introduce KHRONOS: an artificial intelligence framework tailored to the demands of modern computational science and engineering. KHRONOS is designed to operate across the full computing spectrum, from low-power edge devices in robots to exascale supercomputers. KHRONOS represents target fields as a hierarchical composition \cite{saha2021hierarchical} of kernel expansions. In a single hidden layer network, this representation is effectively a Galerkin interpolation built with kernel shape functions, somewhat akin to an interpolating neural network (INN) \cite{Park2024}. In the subsequent sections, this article will discuss KHRONOS's architecture and its application to three major classes of problems - \textit{model-free}: in a pure data-driven environment, \textit{model-based}: where there is a high-dimensional partial differential equation (PDE) to be solved, and \textit{model-inverse}: where output-to-input mappings is needed from the forward mapping. 

\begin{figure}
    \centering
    \includegraphics[width=\linewidth]{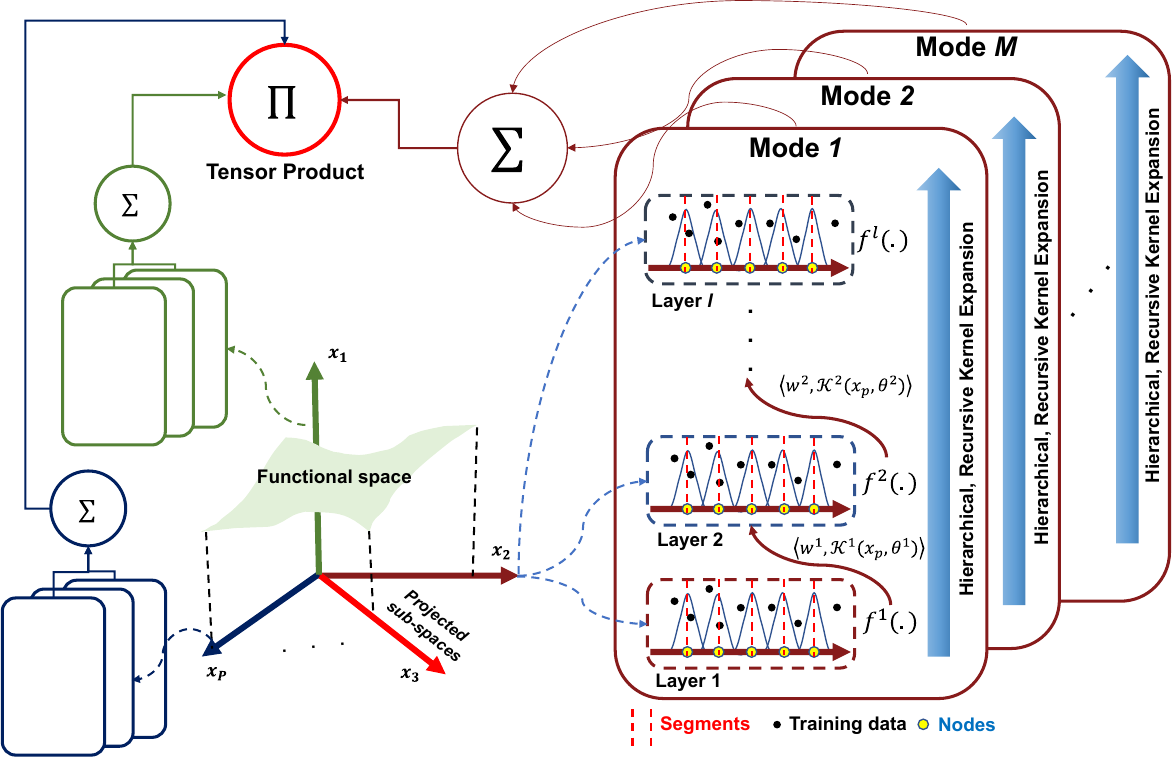}
    \caption{Schematic of KHRONOS's architecture. Each input feature $x_p$ is mapped via a kernel expansion (layers 1-$L$) defined on a small number of nodes (yellow) within each segment. Per-dimension feature vectors are projected by learned weights $w$ and then combined via a tensor product ($\prod$) to form each mode. Finally, $M$ such modes are summed ($\sum$) to yield the surrogate.}
    \label{fig:KHRONOS}
\end{figure}

\section{Methodology of KHRONOS}

\subsection{Architecture}
Figure \ref{fig:KHRONOS} illustrates KHRONOS's core architecture. KHRONOS approximates a high-dimensional functional space by projecting onto one-dimensional feature subspaces. Each input feature space is partitioned into $N^e_p$ segments, inducing a knot vector $\{\theta_i\}_{i=1}^{N_p^e+1}$. These knots are used to construct a finite set of kernels over the space. A key example used throughout this work is that of second-order (quadratic) B-spline kernels \cite{piegl1997nurbs}. These are defined over four consecutive knots and are compactly supported on the three inscribed segments. By extending the knot vector by two points beyond the domain at each end (bringing us to to $N_p^e+5$ total knots/nodes), one obtains $N_p^k=N_p^e+2$ quadratic B-spline basis functions on the domain which satisfy partition-of-unity.

These kernel evaluations are linearly combined via learnable weights into a local feature approximation within each one-dimensional subspace. Each such projection constitutes a single layer, with multiple layers stacked by sending the output of a previous projection as an input into the next. This can be seen as analogous to the successive feature maps in a convolutional neural network (CNN) \cite{lecun1989backpropagation}. Unlike standard networks, however, nonlinearity is inherent to the choice of kernel basis rather than imposed by external activation functions. Each off these one-dimensional feature maps is a mode contribution, with a single mode assembled by their overall product. Several such modes, each with their own learned feature spaces, are then superposed to produce the surrogate output.

\subsection{Mathematical Formulation} 
\subsubsection{Model Ansatz and Forward Propogation}
KHRONOS aims to find a representation $\hat u(x)$ for a target field $u(x)$ over a $P-$ dimensional input feature $x=(x_1,\dots,x_p,\dots,x_P)$. It does this by hierarchically composing per-feature kernel expansions into full-parameter modes. Each feature $x_p$ is first mapped through a sequence of $L$ expansion layers. The feature map of the $l$-th layer for parameter $p$ is denoted,
\begin{align}
    f^{(l)}_{p} = \mathcal{K}(f^{(l-1)}_{p}, \theta^{(l)}_{p}),
\end{align}
where $\theta^{(l)}_{p,i}$ are the kernel's parameters, $f^{(0)}_{p,i}(x_p)=x_p$, and $f_p^{(l-1)}\equiv f_p^{(l-1)}(x_p)$ is the scalar output from the previous expansion layer. Each layer's output is formed by a weighted sum of these kernels, 
\begin{align}
    f_p^{(l)}(x_p)
    &= \sum_{i=1}^{N_{p,l}} w_{p,i}^{(l)}\mathcal{K}\bigl(f_p^{(l-1)}(x_p),\theta_{p,i}^{(l)}\bigr), \\
    &= \sum_{i=1}^{N_{p,l}} w_{p,i}^{(l)}f_{p,i}^{(l)},\\
    &=\langle w^{(l)}_p,f^{(l)}_p\rangle.
\end{align}
After $L$ such compositions, each feature $p$ yields a scalar $f^{(l)}_p(x_p)$. KHRONOS then builds a number of separable modes $M_j(x)$, each learning its own feature-wise layer outputs, $f^{(l)}_{p,j}(x_p)$. Modes are constructed by multiplying across features,
\begin{align}
    M_j(x)=\prod\limits_{p=1}^Pf_{p,j}^{(L)}(x_p).
\end{align}
There are two approaches to training modes,
\begin{enumerate}
    \item Cooperative (joint) learning. 
    The number of modes is predefined as $J$. KHRONOS initializes each with separate parameters and superposes them,
    \begin{align}
        \hat u(x)=\sum\limits_{j=1}^JM_j(x).
    \end{align}
    This superposition is trained in its entirety, with modes thus being trained concurrently.
    
    \item Sequential learning.
    A single mode $M_1(x)$ is initialized and trained. The next mode then seeks to represent the new target field, $u_1(x)=u(x)-M_1(x)$. This is iterated until an acceptable tolerance $\epsilon$ is met, so that the number of modes is  $J_{tol} \;=\; \min\{\,j : \|u - \sum_{i=1}^j M_i\| < \epsilon\}.$
    The surrogate is again the superposition of all of these modes, 
    \begin{align}
        \hat u(x)=\sum\limits_{j=1}^{J_{tol}}M_j(x).
    \end{align}
    
\end{enumerate}

\subsubsection{Loss Functions}

\paragraph{Model-Free} 

A model-free (or supervised learning) approach involves learning from labeled data. KHRONOS, in particular, learns from data structured as $d$-input, scalar-output pairs $\{(x_i,u_i)\}_{i=1}^I,~x_i\in\mathbb{R}^d,~u_i\in\mathbb{R}$. Then for a parametric model $\hat u(x;\theta)$, $\theta$ the model parameters, the model-free, mean-squared error loss is given by
\begin{align}
    L_{mse}=\frac{1}{I}\sum\limits_{i=1}^I(\hat u(x_i;\theta)-u_i)^2.
\end{align}

\paragraph{Model-Based}

For physics-based training (solving), the loss function for a space-time-parameter can be constructed in two ways: a) using collocation-based method akin to PINNs \cite{raissi2019physics}, or b) using Galerkin-like weak formulation. The space-time-parameter is defined over $x\in\Omega$, enclosed by a boundary $\partial\Omega,$ with time $t\in[0,T]$, and parameters $d_1,\dots,d_\rho\in\mathcal{P}\subset\mathbb{R}^{\rho}$. Given a second order spatial differential operator $\mathcal{L}$, first-order boundary differential operator $\mathcal{B}$, source term $f(x,t;d)$ and boundary source term $g(x,t;d)$ the target PDE is defined
\begin{align}
    \label{eq:PDE}
    \partial_tu-\mathcal{L}u&=f\quad\textrm{in }\Omega,\\
    \mathcal{B}u&=g\quad\textrm{on }\partial\Omega, \\
    u&=u_0~~\text{at } t=0. 
\end{align}
For a neural surrogate $\hat u(x,t,d;\theta),$ with \textit{network} parameters $\theta$, residuals are then defined,
\begin{align}
    \label{eq:res}
    r_\Omega &=\partial\hat u-\mathcal{L}\hat u-f, \\
    r_{\partial \Omega} &=-\mathcal{B}\hat u-g.
\end{align}

With given hyperparameters $\alpha_\Omega, \alpha_{\partial\Omega},$ and residuals defined in \ref{eq:res} a strong formulation loss function can then be constructed,
\begin{align}
    L_{strong}(\theta)=\frac{\alpha_{\Omega}}{N_\Omega}\sum\limits_{k,n,l}r_\Omega(x_k,t_n,d_l;\theta)^2 + \frac{\alpha_{\partial\Omega}}{N_{\partial\Omega}}\sum\limits_{b,n,l}r_{\partial\Omega}(x_b,t_n,d_l;\theta)^2,
\end{align}
where $\{x_k\}_{k=1}^{N_\Omega}\subset\Omega,\{x_b\}_{b=1}^{N_{\partial\Omega}}\subset\partial\Omega,\{t_n\}_{n=1}^{N_t}\subset[0,T]$ and $\{d_l\}_{l=1}^{N_d}\subset\mathcal{P}$.
This is a collocation loss, evaluated at $N_\Omega$ interior points and $N_{\partial\Omega}$ boundary points. Collocation based approaches can face sensitivity issues, where careful sampling is required to avoid spiky errors between points. Further, the surrogate is required to be sufficiently smooth in order for $\mathcal{L}u$ to be well defined at collocation points. 

A Galerkin weak formulation has less stringent smoothness requirements. Namely, the spatial requirement is $\hat u(\cdot,t,d) \in H^1(\Omega),\forall^\infty t\in[0,T]$. Then, the bilinear form $a(\hat u,v)=\int_\Omega a\nabla \hat u\cdot\nabla vdx$ is well defined $\forall v\in H^1(\Omega).$ The temporal requirement is that for almost every $x\in\Omega$ and $d\in\mathcal{P},$ $\hat u(x, \cdot ,d)\in L^2(0, T).$ Thus, the choice of second-order splines (or any kernel in $H^1(\Omega)\times L^2(0,T)$) guarantees existence and uniqueness of solutions for a sufficiently regular and coercive operator $\mathcal{L}.$

The weak formulation of \eqref{eq:PDE} is defined
\begin{align}
    \int_0^T\int_\Omega\partial_tuv~dx~dt-\int_0^T\int_\Omega v\mathcal{L}u~dx~dt=\int_0^T\int_\Omega fv~dx~dt,\forall v\in V(\Omega)\times L^2(0,T),
\end{align}
with test space $V\subseteq H^1(\Omega)$ depending on the boundary conditions. Given a test function $\hat u\in H^1(\Omega)\times L^2(0,1),$ the associated weak residual is defined as,
\begin{align}
    \mathcal{R}\{\hat u,v\}=\int_0^T\int_\Omega v\partial_t\hat u-v\mathcal{L}\hat u-fv~dx~dt,~\forall v\in V(\Omega)\times L^2(0,T).
\end{align}
For training, a weak residual loss is then defined as,
\begin{align}
    L_{weak}(\theta)=\sum\limits_{j}(\mathcal{R}\{\hat u,v_j\})^2,
\end{align}
for a finite set of test functions $\{v_j\}\subset V(\Omega)\times L^2(0,T).$ Or for a collocation-based loss,
\begin{align}
    L_{weak}(\theta)=\int_0^T\int_\Omega(\partial_t\hat u-\mathcal{L}\hat u -f)^2v^2~dx~dt,
\end{align}
with a fixed choice for $v$ - typically $v\equiv1$.

For a linear, symmetric, and coercive $\mathcal{L}$ -such as the Laplacian $-\Delta$ used as an example in Section \ref{sec:modelbased} - we may equivalently minimize an energy-based loss.

\paragraph{Separable Integration}
Taking the 2D Poisson equation, with homogeneous Dirichlet boundary conditions, on $[0,1]^2$,
\begin{align}
    -\nabla^2u=f,
\end{align}
as an example, the separable integration approach is laid out. While the following derivation is specific to that example, the core principles are readily extended to different boundary conditions, different differential operators and even different integral forms (i.e. general weak formulations). First, note that the energy functional to minimize is given by
\begin{align}
    \varepsilon(u)=\int_0^1\int_0^1\left(\frac12|\nabla u|^2-fu\right)~dxdy, 
\end{align}
so that,
\begin{align}
    V(u)&=\int_0^1\int_0^1\frac12|\nabla u|^2~dxdy,\\
    U(u)&=\int_0^1\int_0^1fu~dxdy.
\end{align}
Consider KHRONOS's ansatz of 
\begin{align}
    \hat u = \sum\limits_{m=1}^Mg_m(x)h_m(y),
\end{align}
and $f(x,y)=\sum_{i=1}^Nf^x_i(x)f^y_i(y)$, derived either analytically or fitted numerically. Then, 
\begin{align}
V&=\frac12\left(\sum\limits_{m=1}^Mg'_m(x)h_m(y)\right)^2+\frac12\left(\sum\limits_{m=1}^Mg_m(x)h'_m(y)\right)^2,\\
\int_0^1\int_0^1V~dxdy&=\frac12\sum\limits_{i=1}^M\sum\limits_{j=1}^M\left(\int_0^1g'_i(x)g'_j(x)~dx\int_0^1h_i(y)h_j(y)~dy\right)+\dots,
\end{align}
\begin{align*}
    \dots +  \frac12\sum\limits_{i=1}^M\sum\limits_{j=1}^M\left(\int_0^1g_i(x)g_j(x)~dx\int_0^1h'_i(y)h'_j(y)~dy\right).
\end{align*}
hence,
\begin{align}
    \int_0^1\int_0^1V~dxdy=\frac12(trace(G'^TH)+trace(H'^TG)).
\end{align}
Here, $G,G',H,H'$ are the Gram matrices defined by,
\begin{align}
    \{G\}_{i,j}&=\langle g_i(x),g_j(x)\rangle_{L_2},\\\{G'\}_{i,j}&=\langle g'_i(x),g'_j(x)\rangle_{L_2},\\\{H\}_{i,j}&=\langle h_i(x),h_j(x)\rangle_{L_2},\\\{H'\}_{i,j}&=\langle h'_i(x),h'_j(x)\rangle_{L_2}.
\end{align}
Similarly,
\begin{align}
    U&=\sum_{m=1}^Mg_m(x)h_m(y)\sum_{i=1}^Nf^x_i(x)f^y_i(y), \\
    \int_0^1\int_0^1U~dxdy&=\sum_{m=1}^M\sum_{i=1}^N\langle g_m(x),f^x_i(x)\rangle_{L_2}\langle h_m(y),f^y_i(y)\rangle_{L_2}
\end{align}
Then,
\begin{align}
    \int_0^1\int_0^1U~dxdy=trace(A^TB),
\end{align}
with $\{A\}_{m,i}=\langle g_m(x),f^x_i(x)\rangle_{L_2}$ and $\{B\}_{m,i}=\langle h_m(y),f^y_i(y)\rangle_{L_2}$. Hence, the energy functional can be written
\begin{align}
    \varepsilon(\hat u)=\frac12(trace(G'^TH)+trace(H'^TG))-trace(A^TB).
\end{align}
Each of the $L_2(0,1)$-norm evaluations get broken down further into a sum of integrals over each of the $n_e$ elements. Each of these sub-integrals is evaluated by Gauss-Legendre quadrature at $n_\textrm{gauss}$ points, using automatic differentiation for $g'$ and $h'$ contributions. Overall, this pipeline reduces a costly and less accurate $O(n^2)$ integral to one of $O(n_\textrm{gauss}n_e(2M^2+MN))$, where $n_\textrm{gauss},n_e,M,N\ll n$. This reduction becomes only more pronounced in higher-dimensional PDEs.
\paragraph{Mixed Models}
In practice, it is possible to construct a loss function as a combination of model-free and model-based terms. A common choice is to take $\alpha_{data}, \alpha_{model}$, and write,
\begin{align}
    L_{mixed}(\theta)=\alpha_{data}L_{mse}(\theta) + \alpha_{model}L_{weak}(\theta).
\end{align}
Such a formulation is useful in settings with limited data and uncertain or partially known physics, or when an empirical-model balance is required.

\subsubsection{Inverse Modeling}

Inverse modeling is the task of inferring unknown parameters from observed outputs, in particular from a learned model. Formally, let $\hat u:\mathcal{X}\rightarrow \mathcal{Y}$ be a learned KHRONOS surrogate that maps inputs $x\in \mathcal{X}$ to outputs $\hat u\in\mathcal{Y}.$ Given some observed outcome $z\in\mathcal{Y},$ an inverse modeling problem seeks an input $\alpha\in\mathcal{X}$ so that,
\begin{align}
    \hat u(\alpha)=z.
\end{align}
With the right choice of kernel, KHRONOS's constructs a continuously differentiable $\hat u$. This allows for gradient-based root-finding or optimization algorithms. One choice investigated in Section \ref{sec:modelinversion} is Gauss-Newton \cite{fletcher2000practical}. If $x_{k}$ is the $k$-th Gauss-Newton iteration, the next iterate $x_{k+1}$ is found by
\begin{align}
    x_{k+1}=x_{k}-\frac{\hat u - z}{|\nabla\hat u|^2}\nabla\hat u,
\end{align}
with $x_0\in\mathcal{X}$ an initial guess.

Gauss-Newton is lightweight, requiring a single forward and gradient evaluation in one update with automatic differentiation. Further, it is \textit{embarrassingly parallel} across different guesses $x_0$, and different targets $z$. This makes it a strong candidate for \textit{batch inversion}; parallel evaluation of initial conditions sampled over $\mathcal{X}$. This allows for entire level set recovery on the order of single milliseconds. This performance brings inverse modeling into real-time, online and high-throughput regimes from what is traditionally an offline process.

\section{Performance Analysis of KHRONOS}

\subsection{Model-Free}

To assess model-free, supervised performance, KHRONOS is compared to some high-performing contemporary models: Random Forest (RF), XGBoost and a multilayer perceptron (MLP). Two benchmark problems are considered: the 8-dimensional borehole function, and a more challenging 20-dimensional Sobol-G function with added artificial noise.
\subsubsection{8-Dimensional Borehole Function}
\begin{figure}
    \centering
    \includegraphics[width=0.75\linewidth]{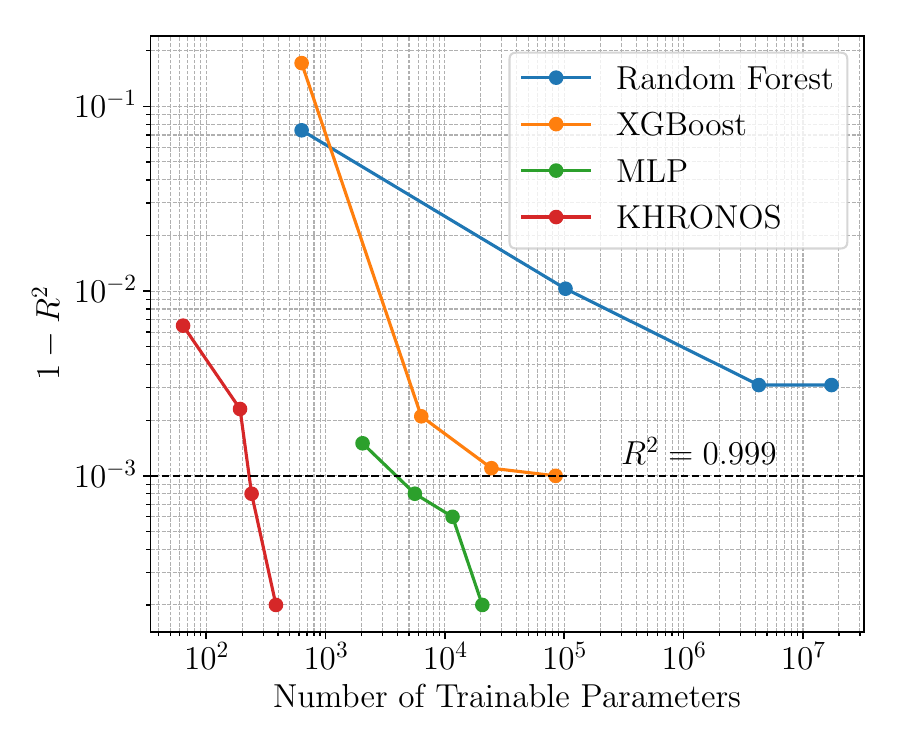}
    \caption{Plot of convergence toward perfect accuracy, $1-R^2\rightarrow0$, as trainable parameters increase for each surrogate model}
    \label{fig:params}
\end{figure}

The toy problem is the 8-dimensional borehole function,
\begin{align}
    \label{eq:borehole}
    u(p)=2\pi p_3(p_4-p_6)\left(\log\left(\frac{p_2}{p_1}\right) \left(1+2\frac{p_7p_3}{\log\left(\frac{p_2}{p_1}\right)p_1^2p_8}+\frac{p_3}{p_5}\right) \right)^{-1},\\
\end{align}
with features,
\begin{align}
    \textrm{Borehole radius ($m$):}\quad p_1&\in[0.05,0.15],\\
    \textrm{Radius of influence ($m$):}\quad p_2&\in[100,50000],\\
    \textrm{Transmissivity of upper aquifer ($m^2/yr$):}\quad p_3&\in[63700,115600],\\
    \textrm{Potentiometric head of upper aquifer ($m$):}\quad p_4&\in[990,1110],\\
    \textrm{Transmissivity of lower aquifer ($m^2/yr$):}\quad p_5&\in[63.1, 116],\\
    \textrm{Potentiometric head of lower aquifer ($m$):}\quad p_6&\in[700,820],\\
    \textrm{Length of borehole ($m$):}\quad p_7&\in[1120,1680],\\
    \textrm{Hydraulic conductivity of borehole ($m/yr$):}\quad p_8&\in[9855,12045].
\end{align}
This function is typical for testing uncertainty quantification and surrogate models. Data, in the form of input-output pairs, is generated by sampling the equation at 100,000 points using Latin Hypercube sampling. This data is normalized, and then split 70/30, train-test. Table \ref{tab:borehole} shows the performance of each models. To provide a consistent saturation point, baseline model complexity (number of trees, maximum depth, number and width of layers) was increased until the model achieved a validation ($R^2$) score of at least 0.999. This parameter saturation is shown in figure \ref{fig:params}. 

\begin{table}[h]
\centering
\caption{Benchmark comparison of surrogate models on the borehole problem \eqref{eq:borehole}, sampled at 100,000 points with a 70/30 train-test split. RF used 100 estimators with a maximum depth of 15, XGBoost had 200 estimators with a max depth of 8, the MLP had 2 hidden layers with widths of 50, and KHRONOS was run with 4 kernels per-dimension and 3 modes.}
\label{tab:borehole}
\begin{tabular}{lcccc}
\toprule
Metric & Random Forest\cite{breiman2001random} & XGBoost \cite{chen2016xgboost} & MLP & KHRONOS \\
\midrule
Trainable parameters & 4,261,376 & 84,600 & 5601 & 240\\
Training time (s)       & 2.8 & 1.5 & 22 & 0.87\\
Inference time (ms)      & 75 & 61 & 0.7 & 0.2\\
Test MSE                & $1.0\times 10^{-4}$ & $3.3\times 10^{-5}$ & $2.8\times 10^{-5}$ & $2.2\times 10^{-5}$ \\
Test R$^2$              & 0.9969 & 0.9990 & 0.9992 & 0.9998\\
\bottomrule
\end{tabular}
\end{table}

RF was unable to achieve the target $R^2$-score, saturating at 0.9969. Remarkably, KHRONOS achieved an $R^2$-score of 0.9935 with as few as 64 trainable parameters. Furthermore, it was the only tested surrogate able to hit the target $R^2$ in under a second. In addition to its low parameter count, this efficiently comes from its computational structure. Whereas MLPs require dense matrix operations of complexity $O(n_{layers}\cdot n_{widths}\cdot n_{widths})$, the dominant cost in KHRONOS is $O(n_{modes}\cdot n_{dim})$ in mode construction. For the borehole problem, this is $O(1)O(10)O(10)$ for an MLP but only $O(1)O(10)$ for KHRONOS.
\subsubsection{High-Dimensional, Noisy Regression: 20D Sobol-G Function}
KHRONOS is next tested under more demanding conditions. A 20-dimensional Sobol-G function,
\begin{align}
    u(p)&=\prod_{i=1}^{20}\frac{|4p_i -2|+a_i}{1+a_i},\\
    a_i&=
    \begin{cases}
        0&\quad\textrm{for }i=1,\dots,5\\
        \frac32&\quad\textrm{for }i=6,\dots,10\\
        4&\quad\textrm{for }i=11,\dots,20
    \end{cases},\\
    p&=[0, 1]^{20},
\end{align}
is chosen to this end. The exact function outputs $u(p)$ are then corrupted with additive Gaussian noise,
\begin{align}
    u_\textrm{noisy}&=u(p)+\epsilon,\\
    \epsilon&\sim \mathcal{N}(0,\sigma^2),
\end{align}
with the standard deviation of noise $\sigma=0.01$. The resulting noisy outputs $u_\textrm{noisy}$ are then sampled at 100,000 points using Latin Hypercube sampling. 

Again, Random Forest, XGBoost, the MLP and KHRONOS are tested to see if it can reach the same target $R^2$-score of 0.999. Table \ref{tab:s1} shows this second comparison of the contemporary data-driven regression models. This time, only KHRONOS retains near-perfect accuracy, (test MSE of $6.8\times 10^{-7}$, test $R^2=0.9994$), while Random Forest, XGBoost and the MLP saturate far below this point.

KHRONOS itself is trained for 1000 epochs with a single mode and 40 elements per dimension, combining to 1560 parameters. Training takes 5.1 seconds with Adam, and the learning rate run on a cosine schedule from 0.15 initially to 0.05. Test-set inference remains sub-millisecond. The MLP was set up using a funnel approach, with 4 hidden layers with respective widths of 128, 64, 32, and 16 neurons. It was trained for 50,000 epochs with Adam, with a fixed learning rate of 0.001. A range of other setups and learning rates were tested, but the provided one performed by far the best. Random Forest and XGBoost had parameters increased until saturation, with RF given 100 estimators and XGBoost given 5000, and set to a maximum depth of 12.

\begin{table}[h]
\centering
\caption{Regression benchmark on a noisy 20D Sobol-G Function, sampled at 100,000 points with LHS.}
\label{tab:s1}
\begin{tabular}{lcccc}
\toprule
Metric & Random Forest \cite{breiman2001random} & XGBoost \cite{chen2016xgboost} & MLP & KHRONOS \\
\midrule
Trainable parameters & 8,849,208 & 239,564 & 13,569 & 1560\\
Training time (s)       & 5.4 & 5.6 & 66 & 5.1\\
Inference time (ms)      & 54 & 61 & 0.3 & 0.9\\
Test MSE& $4.6\times 10^{-4}$ & $3.3\times 10^{-5}$ & $1.4\times 10^{-4}$ & $6.8\times 10^{-7}$ \\
Test $R^2$              & 0.5565 & 0.7312 & 0.8788 & 0.9994\\
\bottomrule
\end{tabular}
\end{table}

\subsection{Model-Based}
\label{sec:modelbased}
\begin{figure}
    \centering
    \includegraphics[width=1.0\linewidth]{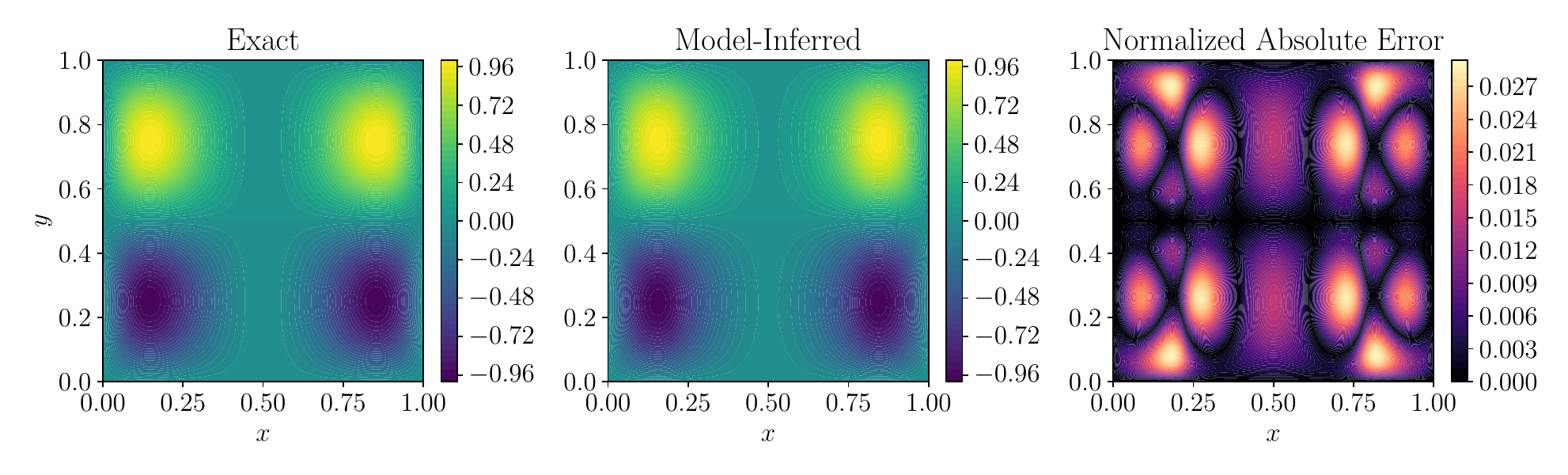}
    \caption{Exact solution, model prediction and the normalized absolute error for a 16-parameter KHRONOS solve}
    \label{fig:poissonex}
\end{figure}

In this section, $L_2^2$ denotes the squared $L_2$-norm over $\Omega=[0,1]^2$, $\|u\|_{L_2(\Omega)}$, and $H_1^2$ denotes the squared $H^1$-seminorm $\|\nabla u\|_{L^2(\Omega)}^2$.
KHRONOS is used as a model-based solver, with a canonical 2D Poisson problem taken as example \cite{Liu2024},
\begin{align}
    \label{eq:poisson}
    -\Delta u&=f\quad\text{in }\Omega,\\
    u&=0\quad\text{on } \partial\Omega,
\end{align}
with $\Omega=[-1,1]^2,~f(x,y)=\pi^2(1+4y^2)\sin(\pi x)\sin(\pi y^2)-2\pi\sin(\pi x)\cos(\pi y^2)$. This system is transformed by $(\tilde x, \tilde y)=2(x,y)-1,$ so that $\tilde \Omega=[0,1]^2,\tilde f(\tilde x, \tilde y)=4f(x,y)$, which admits the same solution $u(\tilde x, \tilde y)$. Equation \eqref{eq:poisson} admits exact solution $u(x,y)=\sin(\pi x)\sin(\pi y^2)$.
The corresponding energy functional is defined
\begin{align}
    \varepsilon(\tilde u)=\int_0^1\int_0^1\left(\frac12|\nabla \tilde u|^2-\tilde f\tilde u\right)~d\tilde x~d\tilde y,
\end{align}
and admits a unique minimizer in $H^1_0(\Omega)$, under standard assumptions on $f\in L^2(\Omega)$. This follows from the Direct Method in the calculus of variations \cite{giusti2003direct}. KHRONOS constructs a kernel-based approximation $\hat u(\tilde x,\tilde y)$ using second-order B-splines. This choice of kernel ensures $\hat u(\tilde x,\tilde y;\theta)\in H_0^1(\Omega)$ and is thus admissible in the variational formulation. It can therefore be trained by minimizing $L(\theta)=\varepsilon(\hat u(\tilde x,\tilde y;\theta))$. In this case, KHRONOS is then, in effect, meshfree, variationally consistent and free of costly matrix operations. Figure \ref{fig:poissonex}
shows an example of a hyper-light 16-parameter KHRONOS solve.

\begin{figure}
    \centering
    \includegraphics[width=\linewidth]{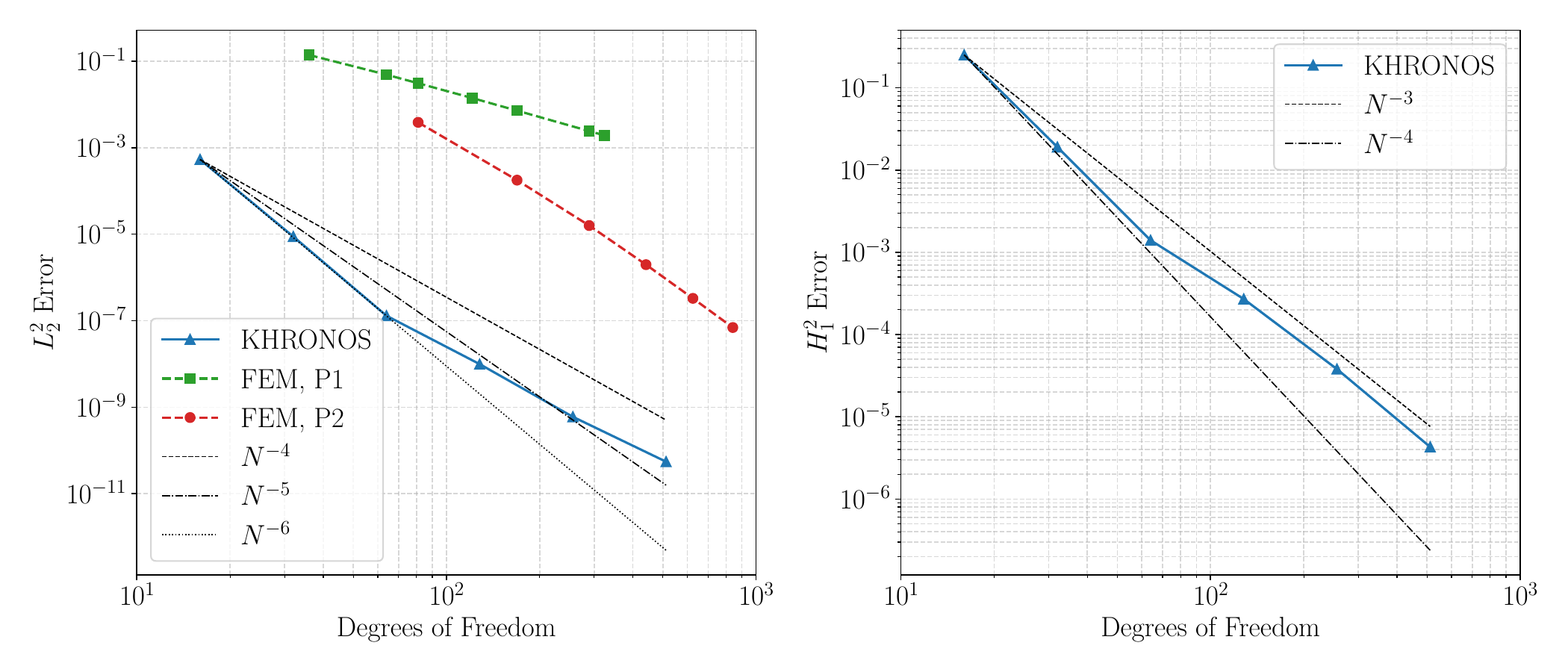}
    \caption{Plot showing $L_2^2$ error and $H_1^2$ error as degrees of freedom increase. The left plot additionally shows the $L^2_2$ errors achieved by P1- and P2 Lagrange element FEM. $N^{-6},N^{-5},N^{-4}$ scaling laws for $L_2^2$ and $N^{-4}, N^{-3}$ scaling laws for $H^2_1$ errors are shown for reference}
    \label{fig:Fig4}
\end{figure}

Table \ref{tab:poisson} summarizes KHRONOS's performance over a range of degrees of freedom (DoFs). Figure \ref{fig:Fig4} presents log-log plots of errors in the square $L_2$-norm and square $H_1$-seminorm. The $L_2^2$ and $H_1^2$ errors exhibit pre-asymptotic empirical scaling laws of DoF$^{-6}$ and DoF$^{-4}$, respectively, before settling into slopes of DoF$^{-4}$ and DoF$^{-3}$. These steep initial slopes may be a characteristic of KHRONOS's automatic r-adaptive process quickly resolving dominant components in the solution. Having been tested on the same problem, KAN \cite{Liu2024} constructed with the same second order b-splines achieves a similar, DoF$^{-4}$ scaling law in the $L_2^2$ error. However, it requires greater than 40 degrees of freedom to achieve the $L_2^2$ error KHRONOS sees with 16 degrees of freedom. Thus while both architectures exhibit similar asymptotic scaling, KHRONOS enjoys a substantial head start. Whereas KAN requires $\sim150$ trainable parameters to drive the $L_2^2$ error down to $10^{-6},$ KHRONOS attains $L_2^2<10^{-8},$ on the same parameter budget - a greater than hundredfold increase in accuracy. With 256 parameters, this greater-than-hundredfold improvement continues. Furthermore, KHRONOS sees a four- or five-order of magnitude improvement on any of the MLP setups tested \cite{Liu2024},

\begin{table}[h]
\centering
\caption{Performance of KHRONOS on the 2D Poisson benchmark with 16, 32, 64, 128, 256 and 512 degrees of freedom, each for 3000 epochs. This test is run on an NVIDIA Ampere A100, 40GB GPU. Inference is run on a $1000\times 1000$ grid.}
\label{tab:poisson}
\begin{tabular}{@{}ccccc@{}}
\toprule
DoF & Epoch Time ($\mu$s) & Inference ($\mu$s) & $L_2^2$ & $H_1^2$ \\
\midrule
16   & 330 & 64 & $5.3 \times 10^{-4}$ & $2.5 \times 10^{-1}$ \\
32   & 516 & 69 & $8.8 \times 10^{-6}$ & $1.9 \times 10^{-2}$ \\
64   & 594 & 71 & $1.3 \times 10^{-7}$ & $1.4 \times 10^{-3}$ \\
128  & 687 & 74 & $9.9 \times 10^{-9}$ & $2.7 \times 10^{-4}$ \\
256  & 804 & 72 & $6.0 \times 10^{-10}$& $3.8 \times 10^{-5}$ \\
512  & 860 & 89 & $5.5 \times 10^{-11}$ & $4.3 \times 10^{-6}$\\
\bottomrule
\end{tabular}
\end{table}

\subsection{Model Inversion}
\label{sec:modelinversion}
In this section, the continuously differentiable nature of the surrogate found by KHRONOS is exploited in order to perform batch model inversion. This is highlighted by the toy problem,
\begin{align}
    u(x,y)=\sin(4\pi x)\sin(2\pi y) +\frac12 \sin(6\pi x)\sin(3\pi y),
\end{align}
on $[0, 1]^2$. KHRONOS is first trained on generated by Latin Hypercube sampling at $n$ points, $n=500,1000,2000,4000,8000$ and $16000$. The goal is then inversion to find the level set $\hat u=0$, via Gauss-Newton for 10 iterations. Table \ref{tab:inversion} reports total latency, per-point latency, convergence failure rate, and RMSE for each batch. As the GPU is saturated with sufficient batch size increases, sub-microsecond per-point inversion times are seen. Failure rates and errors remain steady across the tested batch sizes, highlighting the strength of this divide-and-conquer approach, even in a highly non-convex example.

\begin{table}[t]
\centering
\caption{Batched Newton‐inversion on an A100.  Total elapsed time, per-point latency, failure rate (points with residual >1e-3), and residual}
\label{tab:inversion}
\begin{tabular}{@{}rcccc@{}}
\toprule
Batchsize & Total Time ($ms$) & Time per Point ($\mu s$) & Failure Rate \% & RMSE \\
\midrule
500   & 4.9 & 9.7 & $0.2$ & $1.2\times10^{-3}$\\
1000  & 5.3 & 5.3 & $0.3$ & $1.2\times10^{-3}$\\
2000  & 5.1 & 2.6 & $0.3$ & $1.2\times10^{-3}$\\
4000  & 6.2 & 1.5 & $0.1$ & $1.2\times10^{-3}$\\
8000  & 5.9 & 0.7 & $0.3$ & $1.2\times10^{-3}$\\
16000 & 6.9 & 0.4 & $0.3$ & $1.2\times10^{-3}$\\
\bottomrule
\end{tabular}
\end{table}

\begin{figure}
    \centering
    \includegraphics[width=\linewidth]{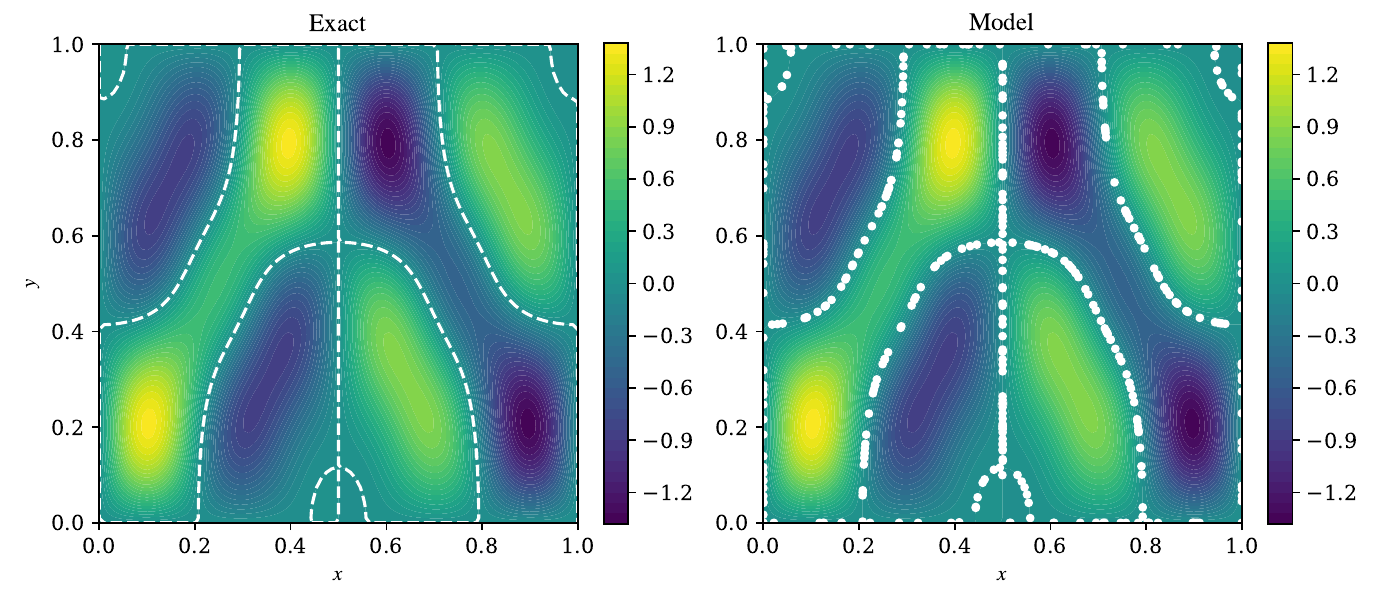}
    \caption{Batch inversion, 400 points, of KHRONOS on a highly non-convex toy example}
    \label{fig:Fig6}
\end{figure}


\section{Conclusions}

This work has presented KHRONOS, a separable, kernel-based surrogate architecture that unifies model-free, model-based and model-inverse learning. Empirical results demonstrated that KHRONOS:
\begin{itemize}
    \item \textbf{Model-free:}  outperforms Random Forest, XGBoost and multilayer perceptron baselines with reduced training times - the only model to achieve a target $R^2$-score in under a second - and one to four orders of magnitude fewer parameters on the 8D borehole benchmark. Furthermore, it achieves $R^2>0.99$ with a remarkably low number of trainable parameters: 64. On the more challenging 20-dimensional noisy Sobol-G benchmark, KHRONOS maintained near-perfect accuracy ($R^2$=0.9994), dramatically surpassing the contemporary models, which saturated at significantly lower performance levels while using more than an order of magnitude more parameters.
    \item \textbf{Model-based:} achieves pre-asymptotic superconvergence $L_2^2\sim$ DoF$^{-6}$, and high-order asymptotic scalings of $L_2^2\sim$ DoF$^{-4}$ and $H_1^2\sim$ DoF$^{-3}$ on a 2D Poisson benchmark, slashing the number of trainable parameters compared to FEM, MLPs and KANs while demonstrating lower $L_2^2$ and $H_1^2$ errors, as well as a single digit second training (up to 2.6s with 512 DoF), and inference times on a $1000\times 1000$ grid in the double digit microseconds (up to 89$\mu$s with 512 DoF).
    \item \textbf{Model-inverse:} enables batched Gauss-Newton inversion for highly nonconvex targets at sub-microsecond-per-sample latency, with robust convergence across thousands of initializations.
\end{itemize}

\section{Limitations}
KHRONOS's current implementation assumes a regular grid over $[0,1]^d$, thus cannot yet handle unstructured meshes or non-rectangular geometries. This restricts its immediate applicability in scenarios requiring complex domain representations, such as CAD geometries. Secondly, the current iteration only uses second-order B-spline basis functions. While these have proven effective thus far, this choice might not be optimal for other problems. Solutions requiring smoothness might suit higher order splines, problems involving discontinuities might benefit from special kernels, and time-dependent problems might benefit from time-history kernels.

While KHRONOS has demonstrated strong performance on canonical regression (8D Borehole, noisy 20D Sobol-G) and PDE (2D Poisson) benchmarks, further validation across a broader spectrum of complex and multidimensional tests is warranted to fully delineate KHRONOS's capabilities. This is especially the case in PDEs, where performance characteristics on more intricate, nonlinear space-time-parameter systems require investigation.
As a novel framework, the development of pre- and post-processing utilities, as well as community-vetted best practices are ongoing processes that would aid wider adoption.

\section{Future Work}

KHRONOS can potentially be extended and applied in many fields of science and engineering, including online monitoring and control of additive manufacturing, inverse design of microstructure, multiscale computation of hierarchical materials systems, and computer vision algorithms for autonomous robotics. Being a kernel-based method, it is natural to apply KHRONOS to image-based problems. Thus far, KHRONOS has shown promise in efficiently learning differentiable image representations. Indeed, preliminary work has demonstrated KHRONOS's strong potential in this domain: an approach using KHRONOS to generate latent-space representations from microstructure images, from which a secondary KHRONOS learns material properties. This framework is therefore also inverse-compatible: one can fix a target property and generate candidate microstructures exhibiting that property.
 
Regarding the separable integration technique for model-based learning, the current formulation assumes a separable source term $f$. A posited approach for handling inseparable source terms is to first approximate them with a separable KHRONOS surrogate. Work in this direction is ongoing.

Finally, continued and more extensive testing against a wider range of contemporary architectures and across a spectrum of benchmarks is required to fully assess KHRONOS's performance characteristics.

\section{Acknowledgments}
S. Saha gratefully acknowledges the start-up fund provided by the by the Kevin T. Crofton Department of Aerospace and Ocean Engineering, Virginia Polytechnic Institute and State University. R. Batley acknowledges the Crofton Fellowship from the Kevin T. Crofton Department of Aerospace and Ocean Engineering, Virginia Polytechnic Institute and State University.

\bibliographystyle{unsrt}
\bibliography{references.bib}

\end{document}